\documentclass[runningheads]{llncs}

\usepackage[super]{nth}
\usepackage{graphicx}
\usepackage[utf8]{inputenc}
\usepackage{amsmath}
\usepackage{bm}
\usepackage{gensymb}
\usepackage{amssymb}
\usepackage{siunitx}
\usepackage{microtype}
\usepackage{xspace}
\usepackage{xcolor}
\usepackage{siunitx}
\usepackage[normalem]{ulem}
\usepackage[hyphens]{url}
\usepackage{multirow}
\usepackage{booktabs}
\usepackage{nth}
\usepackage[pagebackref=true,breaklinks=true,colorlinks,bookmarks=false,citecolor=green!80!black,linkcolor=red!80!black,urlcolor=blue]{hyperref}
\usepackage{cleveref}
\usepackage{subcaption}
\usepackage{tikz}
\usepackage{pgfplots}
\pgfplotsset{compat=newest}
\usepgfplotslibrary{statistics}
\usetikzlibrary{external,patterns,shapes,pgfplots.groupplots,arrows,positioning,fit,backgrounds,calc,spy,decorations.markings}

\usepackage{cite}
\tikzstyle{blockbody} = [rectangle, draw, fill=blue!20,
		text centered, rounded corners,
	minimum height=2em, thick]
\tikzstyle{blockwhite} = [rectangle, draw,
		text centered, rounded corners,
	minimum height=2em, thick]
\tikzstyle{block} = [rectangle, draw=black, fill=blue!20,
		text width=5.5em, text centered, rounded corners,
	minimum height=2em, thick]
\tikzstyle{blockzero} = [rectangle, draw=white, fill=white,
		text=white, text width=5.5em,
		text centered, rounded corners,
	minimum height=2em, thick]

\tikzstyle{blockh} = [rectangle, draw, fill=red!20,
		text width=5.5em, text centered, rounded corners,
	minimum height=2em, thick]

\tikzstyle{blockl} = [rectangle, draw, fill=blue!10,
		text width=5.5em, text centered, rounded corners,
	minimum height=2em, thick]
\tikzstyle{blocklgray} = [rectangle, draw, fill=blue!10, text=gray!70,
		text width=5.5em, text centered, rounded corners,
	minimum height=2em, thick]
	\tikzstyle{blocklnotext} = [rectangle, draw, fill=blue!10, text=blue!10,
		text width=5.5em, text centered, rounded corners,
	minimum height=2em, thick]

\tikzstyle{roundish} = [draw, rectangle, fill=red!20, rounded corners,
text width=5em, text centered, minimum height=2em, thick, inner sep=0em]
\tikzstyle{hhilit} = [draw=black, thick, dotted, inner xsep=0.5em,
inner	ysep=0.5em]
\tikzstyle{newpart}=[align=center,black,thin,draw, rounded	corners,
fill=green!20]
\tikzstyle{newerpart}=[align=center,black,thin,draw, rounded	corners,
fill=orange!20]

\definecolor{sgrey}{HTML}{758188}
\definecolor{slila}{HTML}{712acd}
\definecolor{sgreen}{HTML}{31c54b}
\definecolor{sblue}{HTML}{0f19fb}
\definecolor{sred}{HTML}{f91835}

\crefname{section}{Sec.}{Sections}
\crefname{figure}{Fig.}{Figure}
\crefname{table}{Tab.}{Table}
\crefname{equation}{Equ.}{Equation}
\usepackage[inline]{enumitem}
\setlist*[enumerate]{label=(\arabic*)}

\newcommand{\onedot}{.\xspace}
\newcommand{\etal}[1]{#1~et~al\onedot}

\newcommand{\eg}{e.\,g.,\xspace}

\newcommand{\cf}{cf\onedot}
\newcommand{\ie}{i.\,e.,\xspace}

\newcommand{\aka}{a.\,k.\,a\onedot}
\newcommand{\music}{\textsc{MusicDocs}\xspace}
\newcommand{\icdar}{\textsc{Icdar17}\xspace}
\newcommand{\czby}{\textsc{CzByChron}\xspace}
\newcommand{\musicnsc}{MusicDocs\xspace}
\newcommand{\icdarnsc}{Icdar17\xspace}
\newcommand{\czbynsc}{CzByChron\xspace}

\newcommand{\dist}[1]{d_\text{#1}}

\begin{document}
\title{Re-ranking for Writer Identification and Writer Retrieval}
\author{Simon Jordan\inst{1} \and
Mathias Seuret\inst{1} \and
Pavel Král\inst{2} \and
Ladislav Lenc\inst{2} \and
Jiří Martínek\inst{2} \and
Barbara Wiermann\inst{3} \and
Tobias Schwinger\inst{3} \and
Andreas Maier\inst{1} \and \\
Vincent Christlein\inst{1}
}
\authorrunning{S.\ Jordan et al.}
\institute{Pattern Recognitition Lab, FAU Erlangen-Nuremberg, Erlangen, Germany\\
  \email{simon.simjor.jordan@fau.de}, \email{\{firstname.lastname\}@fau.de} \and 
Dept. of Computer Science and Engineering, University of West Bohemia, Plzeň, Czech Republic\\
    \email{\{pkral,llenc,jimar\}@kiv.zcu.cz} \and
Sächsische Landesbibliothek, Staats- und Universitätsbibliothek Dresden, Dresden, Germany\\
\email{barbara.wiermann@slub-dresden.de}, \email{ortus\_ts@t-online.de}
}

\maketitle              %
\begin{abstract}
Automatic writer identification is a common problem in document analysis. 
State-of-the-art methods typically focus on the feature extraction step with traditional or deep-learning-based techniques. 
In retrieval problems, re-ranking is a commonly used technique to improve the results.
Re-ranking refines an initial ranking result by using the knowledge contained in the ranked result, \eg by exploiting nearest neighbor relations.
To the best of our knowledge, re-ranking has not been used for writer identification/retrieval. 
A possible reason might be that publicly available benchmark datasets contain only few samples per writer which makes a re-ranking less promising.
We show that a re-ranking step based on $k$-reciprocal nearest neighbor relationships is advantageous for writer identification, even if only a few samples per writer are available.
We use these reciprocal relationships in two ways: encode them into new vectors, as originally proposed, or integrate them in terms of query-expansion.
We show that both techniques outperform the baseline results in terms of mAP on three writer identification datasets.
\end{abstract}
\section{Introduction}

In the past decades, vast amounts of historical documents have been digitized and made publicly available. Prominent providers, among others, are the British Library\footnote{http://www.bl.uk/manuscripts/} or the `Zentral- und Landesbibliothek Berlin'.\footnote{https://digital.zlb.de}
Searchable catalogues and archives can ease historical investigations, but considering the tremendous amount of available documents, manual examination by historical experts is no longer feasible. Hence automatic systems, which allow for investigations on a greater scale, are desired. Since the historical documents are provided as digital images, they are well suited for machine learning methods. 
In this work, we will look at a particular case, namely writer retrieval, which is particularly useful in the field of digital humanities, as the identities of writers of historical documents are frequently unknown.

This task is challenging to most robust classification systems, suitable for identification tasks, since they are only applicable in supervised scenarios. 
Therefore, those are not well suited for our particular case, where we would like to also identify \emph{new} writers, who are unknown to the system during the training stage, \aka zero-shot classification. 
Hence, we focus on robust writer retrieval, which does not directly identify the scribe of a document, but provides a ranked list consisting of the most similar, already known writers.

Our main contribution is the introduction of a re-ranking step, which aims to improve the overall result by refining the initial ranking in an unsupervised manner. 
Therefore, we propose two different re-ranking methods, which we adapt to our case. 
First, we employ a well-known re-ranking method based on the Jaccard distance and $k$-reciprocal neighbors. 
These neighbors are furthermore exploited in a novel re-ranking technique through combination with query expansion. 
We show that even retrieval tasks with low gallery size can benefit from re-ranking methods.
Another contribution consists in the newly created \czby corpus which will be made freely available for research purposes.

The structure of this work is as follows. 
After discussing related work in the field of writer identification/retrieval and re-ranking in \cref{sec:related_work}, we describe our writer-retrieval pipeline in \cref{sec:methodology}. There, we also propose two different re-ranking methods, which can theoretically be applied to most information retrieval tasks, but which we adapted to better fit our particular problem.
Finally, we present a detailed evaluation on these methods on three different diverse datasets (\music, \czby, \icdar) in \cref{sec:evaluation}.  

\section{Related Work} %
\label{sec:related_work}

\subsection{Writer Identification} 
Offline text-independent writer identification and retrieval methods
can be grouped into codebook-based methods and codebook-free methods.
In codebook-based methods, a codebook is computed that serves as background model. 
Popular codebooks are based on GMMs~\cite{Fiel13,Christlein15GCPR,christlein17writer} or $k$-means~\cite{christlein17unsup,Christlein18DAS}. 
Such a model is then used to compute statistics that form the global descriptor, \eg first order statistics are computed in VLAD encoding~\cite{Jegou12ALI}.

Conversely, codebook-free methods compute a global image descriptor directly from the handwriting, such as the width of the ink trace~\cite{Brink12} or the so called Hinge descriptor~\cite{Bulacu07} which computes different angle combinations evaluated at the contour. 

Recent methods rely on deep learning, which can still fall in both groups depending on their usage. 
However, apart from some exceptions~\cite{Tang16}, the most methods use Convolutional Neural Networks (CNN) to compute strong local descriptors by using the activations of a layer as features~\cite{Fiel13,Christlein15GCPR,christlein17unsup,Keglevic18} followed by an encoding method to compute a global descriptor.  
In this work, we rely on both, traditional SIFT~\cite{lowe04distinctive} descriptors and VLAD-encoded CNN activation features provided by the work of \etal{Christlein}~\cite{christlein17unsup}, while we evaluate different methods to improve the initial rankings.

\subsection{Re-ranking}

Since we tackle a writer retrieval problem, research in the field of information retrieval is also of interest. Relevant work in this field was proposed by Arandjelovi{\'c} and Zisserman~\cite{arandjelovic12three}, and \etal{Chum}~\cite{chum07total}, both in general looking at specific Query Expansion (QE) related methods. For a detailed survey about automatic QE in information retrieval see Carpineto and Romano~\cite{carpineto12survey}.

Arandjelović and Zisserman~\cite{arandjelovic12three} propose discriminative QE, which aims to consider 'negative' data, provided by the bottom of the initial ranking. While positive data can be added to the model by averaging (AQE), incorporating negative data is a more complex process. 
Beside QE, special nearest neighbor (NN) relationships, provided by an initial retrieval result, can be used to improve performance. In the work of \etal{Qin}~\cite{qin11hello}, unidirectional NN-graphs are used to treat different parts of the ranked retrieval list with different distance measures. Since in our case however, the top of the list is of special interest, the re-ranking method proposed by \etal{Zhong}~\cite{zhong17re} is better suited, which we were able to adapt to our scenario, and will describe in detail in section \ref{sec:jaccard}.

\section{Methodology}
\label{sec:methodology}

\subsection{Pipeline}
\begin{figure}[t]
    \centering
	\begin{tikzpicture}%
		\sf\small

		\node[](in){
		Input
		};
		\node[block,text width=5.5em,right=0.5cm of in](sampling){
			Sampling
		};		
		\node[block,text width=9.2em,right=0.5cm of sampling](fe){
			Local Descriptors
		};
		\node[block,text width=5em,right=0.5cm of fe](enc){
			Encoding
		};

		\node[blockl,text width=5.5em, below=1cm of sampling.center] (b1)
		{
			\begin{tabular}{c}
				Keypoints\\
			\end{tabular}
		};
		\node[blockl,text width=9.2em, below=1cm of fe.center] (b2)
		{
			\begin{tabular}{c}
			    SIFT\\
				CNN-based\\				
			\end{tabular}
		};
		\node[blockl,text width=5em, below=1cm of enc.center] (b3)
		{
			\begin{tabular}{c}
				GVR\\
				VLAD\\
			\end{tabular}
		};
		
		\node[right=0.5cm of enc](out){};
		
		\draw[->](in)--(sampling);
		\draw[->](sampling)--(fe);
		\draw[->](fe)--(enc);
		\draw[->](enc)--(out);
		\draw[-](sampling)--(b1);
		\draw[-](fe)--(b2);
		\draw[-](enc)--(b3);
		
\end{tikzpicture}
    \caption{Processing pipeline to obtain a global descriptor from an input sample.}
    \label{fig:feature_creation}
\end{figure}
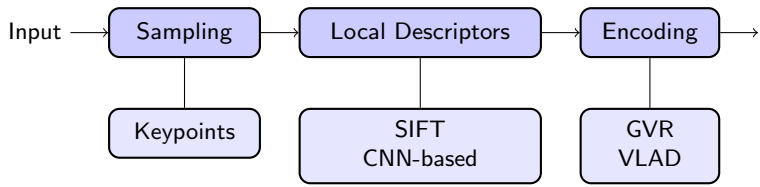

This paper focuses on the re-ranking of given embeddings. 
These embeddings were created using two different instances of the same pipeline, \cf \cref{fig:feature_creation}.
In case of the \icdar dataset, we rely on the embeddings of \etal{Christlein}~\cite{vincentDiss}, where the embeddings are computed as follows. 
First, local descriptors are computed by means of a deep neural network and evaluated at patches located at SIFT keypoints~\cite{lowe04distinctive}. 
The local descriptors are aggregated and encoded using GVR encoding~\cite{Zhou10NGV} and normalized using power normalization~\cite{Perronnin10IFK}, \ie each element of the encoding vector is normalized by applying the square-root to it.  
Afterwards, a feature transformation based on Exemplar SVMs (ESVM-FE)~\cite{christlein17writer} is applied.
Therefore, the embeddings are transformed to a new space using each sample of the test set as single positive sample and all samples of the (disjoint) training set as negatives. 
The new features are the coefficients (weight vector) of the fitted linear SVM, normalized such that the $\ell^2$ norm of this representation is~1. 

For the newly created datasets \czby and \music, we are using in essence a similar pipeline, but with several modifications.
Instead of binarizing the image, we are using the contour image, acquired with the Canny edge detector~\cite{Canny}. 
The hysteresis thresholds are determined by using a Gaussian mixture model with two mixtures~\cite{Wahlberg15}.
Using this image as input, we extract \num{10000} SIFT descriptors computed from SIFT keypoints. 
Afterwards, the SIFT descriptors are Dirichlet normalized~\cite{Kobayashi14} and a PCA-whitening is applied to counter over-frequent bins and correlation, respectively. 
The local descriptors are encoded using VLAD encoding in combination with power normalization and Generalized Max Pooling~\cite{Murray16}, which showed to be beneficial for writer identification~\cite{Christlein18DAS}. 
Afterwards, the global vector representations are once more PCA-whitened and transformed via ESVM-FE.

\subsection{Re-ranking}

Once an initial ranking is obtained by the pipeline described above, it is desirable to improve this result by using knowledge included in the ranking itself. This is possible with both Query Expansion and Re-ranking methods. 

Query Expansion tries to ease the problem by expanding the original model with information provided by the top-ranked samples. Relevant work in this field was published by Arandjelović and Zisserman~\cite{arandjelovic12three} and \etal{Chum}~\cite{chum07total, chum11total}. Re-ranking employs reciprocal rank order knowledge, and thus looks at whether or not two samples are ranked within the top-$n$ results of each other. 

In this section, we first look into a recent re-ranking method proposed by \etal{Zhong}~\cite{zhong17re} using Reciprocal Nearest Neighbours (rNN) and the Jaccard distance. Then, we present our own query expansion approaches, which extend the original ESVM and use rNN to overcome the lack of spatial verification when using our features.

\subsubsection{Jaccard Distance Re-ranking}\label{sec:jaccard}

The problem tackled by \etal{Zhong}~\cite{zhong17re} is, like ours, an image based retrieval task. Instead of scanned documents from different writers, \etal{Zhong} aim to re-identify a person (query) based on images stored in a database. For a given query image, the database is searched in order to return a list of the most similar images (persons). This initial ranking is then improved using the following two techniques.

\paragraph{$k$-reciprocal Nearest Neighbours}\label{krNN}
While it is possible to directly work with the initial ranking for computing the re-ranking, false matches are likely to appear and their presence has a negative effect on the re-ranking~\cite{zhong17re}. Thus \etal{Zhong} make use of the more constrained reciprocal nearest-neighbour (rNN) relationship instead. This relationship is defined as follows: two samples $\mathbf{q}$ and $\mathbf{t}$ are considered $k$-rNN, when both appear within the $k$ highest ranked samples of each other. While the rNN relationship itself is symmetric, the resulting sets of $k$-rNNs are not. Given a query $\mathbf{q}$ with initial ranking result $\pi_q$, the set $\mathcal{Q}_k$ of its $k$-rNNs holds only those of the first $k$ samples in $\pi_q$, which have sample $\mathbf{q}$ appear within their highest $k$ ranked samples. Thus the number of samples in the set $\mathcal{Q}_k$ depends on the query $\mathbf{q}$ and the value of $k$. It is possible, especially for low values of $k$, that a sample has no $k$-rNNs at all.

\paragraph{Jaccard Distance}
Two such sets, $\mathcal{Q}_k$ and $\mathcal{T}_k$, can then be used to define a similarity between the corresponding samples $\mathbf{q}$ and $\mathbf{t}$. For this purpose, we follow the approach from \etal{Zhong}~\cite{zhong17re} and use the Jaccard distance, which gets a lower value, the more samples are shared between both sets, and is maximal if no sample is shared at all. It is given as:

\begin{equation}
\dist{Jaccard} \left( \mathbf{q}, \mathbf{t} \right) = 1 - \frac{\vert \mathcal{Q}_k \cap \mathcal{T}_k \vert}{\vert \mathcal{Q}_k \cup \mathcal{T}_k \vert} \,.
\label{eq:jaccard}
\end{equation}

It is convenient to encode the set information into binary vectors $\boldsymbol{\eta}$, with $\boldsymbol{\eta}_i = 1$ when sample $\mathbf{x}_i$ is part of the corresponding $k$-rNNs set and $\boldsymbol{\eta}_i = 0$ otherwise. This way it is also possible to incorporate the distance $d\left( \mathbf{q}, \mathbf{x}_i \right)$ between the samples and the query, by changing the set vector accordingly, 
\begin{equation}
\boldsymbol{\eta}_i = \begin{cases}
 e^{-d\left( \mathbf{q}, \mathbf{x}_i \right) } & \text{if } \mathbf{x}_i \in \mathcal{Q}_k \\
0 & \text{otherwise}
\end{cases} \,.
\label{eq:jaccvec}
\end{equation}

Then, the Jaccard distance of the sets $\mathcal{Q}_k$ and $\mathcal{T}_k$ with set-vectors $\boldsymbol{\eta}^q$ and $\boldsymbol{\eta}^t$ can be calculated as
\begin{equation}
\dist{Jaccard} \left( \mathbf{q}, \mathbf{t} \right) = 1 - \frac{ \sum_{i=1}^{N} \min \left( \boldsymbol{\eta}_i^q ,  \boldsymbol{\eta}_i^t \right) }
{\sum_{i=1}^{N} \max \left( \boldsymbol{\eta}_i^q ,  \boldsymbol{\eta}_i^t \right)} \,,
\label{eq:jacfinal}
\end{equation}
where $N$ is the total number of vector elements, $\min \left( \boldsymbol{\eta}_i^q ,  \boldsymbol{\eta}_i^t \right)$ the minimum of the respective vector elements at position $i$ and $\max \left( \boldsymbol{\eta}_i^q ,  \boldsymbol{\eta}_i^t \right)$ the maximum of those elements~\cite{zhong17re}. The minimum Jaccard distance, for two equal set-vectors $\boldsymbol{\eta}^q =  \boldsymbol{\eta}^t$ is $\dist{Jaccard} = 1 - 1 = 0$. With the maximum distance being $1$, the range of the Jaccard distance is $[0, 1]$. 
While the initial distance between the samples is already incorporated in the vector entries, see Eq.~\eqref{eq:jaccvec}, and thus does influence the Jaccard distance, \etal{Zhong}~\cite{zhong17re} re-rank the samples based on a weighted sum of the original distance and the Jaccard distance, resulting in the following final distance $d_{\text{J}\lambda}$,
\begin{equation}\label{eq:finaljaccard}
d_{\text{J}\lambda}\left( \mathbf{q}, \mathbf{t} \right) = \left( 1 - \lambda \right) \dist{Jaccard}\left( \mathbf{q}, \mathbf{t} \right) + \lambda d\left( \mathbf{q}, \mathbf{t} \right) \,,
\end{equation}
with $d\left( \mathbf{q}, \mathbf{t} \right)$ being the distance between $\mathbf{q}$ and $\mathbf{t}$ the initial ranking was based on. The weight term $\lambda$ lies within $[0, 1]$. If $\lambda$ is set to $1$, only the original distance will be considered and the initial ranking will stay unchanged. With $\lambda = 0$, the original distance is ignored and the new ranking is solely based on the Jaccard distance. 

Note that our implementation differs slightly from \eqref{eq:jacfinal}, since we add a constant of $\varepsilon = 10^{-8}$ to the denominator in order to enhance numerical stability in case the query $\mathbf{q}$ and test sample $\mathbf{t}$ share no rNN at all. By adding a constant small $\varepsilon$, we ensure %
to avoid a division by zero. 
We observe such cases as we use a lower value of $k$ and work on other data than \etal{Zhong}~\cite{zhong17re}.

\subsubsection{Query Expansion}
\label{sec:QE}

A more common approach to boost the performance in information retrieval systems is automatic query expansion (QE). With an initial ranking given, the original query gets expanded, using information provided by the top-$n$ highest ranked samples. Since those samples can also include false matches, a spatial verification step is often applied to ensure that only true matches are used. Such verification needs to be obtained by employing a different measure than the one used for the initial ranking. For image based tasks, typically it is also important to consider the region of interest (ROI) in the retrieved images, where a single image can contain several query objects. 
In our case, the features do not encode any global spatial information, so we do not have a meaningful spatial verification procedure at hands. With the lack of spatial information, it is also not possible to make use of meaningful regions of interest. 

\etal{Chum}~\cite{chum07total} proposed the following QE approach. After the spatially verified top-$n$ samples $\mathcal{F}$ from the initial ranking have been obtained, a new query sample can be formed using average query expansion (AQE),
\begin{equation}
	\mathbf{q}_{\text{avg}} = \frac{1}{n + 1} \left( \mathbf{q}_0  + \sum_{i = 1}^n \mathbf{f}_i \right) \,,
\end{equation}
with $\mathbf{f}_i$ being the $i$th sample in $\mathcal{F}$ and $n$ being the
total number of samples $\vert \mathcal{F} \vert$. To obtain the improved
ranking, the database gets re-queried using $\mathbf{q_{\text{avg}}}$ instead of $\mathbf{q}$. 

\subsubsection{Pair \& Triple SVM}
\label{sec:pair_triple}
When we consider our feature vectors, obtained using ESVMs, as feature encoders, another way of query expansion is more straightforward. Instead of forming a new query feature vector by averaging, we can expand our model, the ESVM, by adding more samples to the positive set. When a set of suitable \emph{friend} samples $\mathcal{F}$ is given, we can extend the original positive set from the ESVM, which only holds the query sample $\mathbf{q}$, and obtain a new SVM. This new SVM, like the original ESVM, provides a weight vector which allows us to use it as a feature encoder like before. Note that the number of positive samples does not affect the dimension of the weight vector. The cosine distance may again be applied as a similarity measure. Beside the information we provide for the SVM to correctly classify the positive set, nothing has changed. Thus, adding more positive samples will affect the decision boundary and thus the weight vector, but not how we define a similarity based on it. 

While this approach may benefit from larger \emph{friend} sets $\mathcal{F}$, we choose to focus on pair and triple SVMs, by using the most suited \emph{friend} sample, and two most suited \emph{friend} samples, respectively. This way, our QE-re-ranking methods are not limited to scenarios with large gallery sizes $n_G$. In case of the \icdar dataset for example, only four true matches per query are present within the whole test set, and thus using more than two additional positive samples increases the risk of adding a false match to the positive set tremendously. 

Also since we lack any form of spatial verification, we still risk to impair our
model when the selected \emph{friend} samples contain a false match. To minimize
this risk, we use the more constrained $k$-rNNs of the query $\mathbf{q}$ as
basis for our \emph{friend} set $\mathcal{F}_{\text{rNN}}$ which we use for rNN-SVM. This however introduces a new hyper-parameter $k$, which needs to be optimized. Since the intention of the \emph{friend} set $\mathcal{F}$ is to hold only true matches, it follows naturally that optimal values for $k$ may lie within the range of the gallery size $n_G$.

In case of very small gallery sizes, small values for $k$ may be chosen, which
then affects the size of $\mathcal{F}_{\text{rNN}}$. In some cases,
$\mathcal{F}_{\text{rNN}}$ may only consist of a single sample, and thus only a
pair SVM is possible. If $\mathcal{F}_{\text{rNN}}$ holds more than two samples,
those with the smallest original distance to $\mathbf{q}$ are selected for the
triple SVM. In cases where $\mathcal{F}_{\text{rNN}}$ holds no samples at all, we will stick to the original ESVM-FE method; which will result in the original ranking. That way, our proposed QE approach is dynamically adapted to the information provided by the initial ranking and we minimize the risk to end up with a worse result compared to what was our initial guess.

\section{Evaluation}
\label{sec:evaluation}

\subsection{Error Metrics}
\paragraph{Mean Average Precision} (mAP)
\def\AvePQ{\operatorname{AveP}(q)}
\def\relK{\operatorname{rel}(k)}
is calculated as the mean over all examined queries $q$ of set $\mathcal{Q}$:
\begin{equation}
\mathrm{mAP} = \frac{\sum_{q \in \mathcal{Q}} \AvePQ}{\vert \mathcal{Q} \vert} \,,
\end{equation}
with $\AvePQ$ being the average precision of query $q$, given as:
\begin{equation}
\AvePQ = \frac{\sum_{k = 1}^{n} \left( P(k) \times \relK \right) } {\text{number\ of\ relevant\ documents}} \,,
\end{equation}
with $n$ being the total number of retrieved documents, $\relK$ being an indicator function, which is $1$ if the retrieved sample at rank $k$ is relevant (from the same writer as the query) and $0$ otherwise, and $P(k)$ being the precision at position $k$. 

\paragraph{Top-$1$.}

Top-$1$ is equal to precision at $1$ ($P@1$) and thus tells us the percentage of cases when we retrieved a relevant document as highest ranked ($k = 1$) document. Generally speaking, it represents the probability that the highest ranked sample is a true match (from the same writer as the query sample $q$). This makes the Top-$1$ score very informative for our Pair- and Triple-SVM methods, where we rely on a true match as highest ranked result to expand our model in the re-ranking step.

\paragraph{Hard-$k$ \& Soft-$k$.} 

Hard-$k$ represents the probability of retrieving only true matches within the first $k$ retrieved samples. Thus, with only four true matches possible in total, the Hard-$k$ probability for $k \geq 5$ is $0\%$ in case of the \icdar dataset. Unlike the Hard-$k$ metric, Soft-$k$ is increasing for increasing $k$, since it represents the probability of retrieving \textit{at least one} relevant sample within the first $k$ retrieved samples. We chose $k = 5$ and $k = 10$ for our Soft-$k$ measurements and $k = 2, 3, 4$ for Hard-$k$. 

\subsection{Datasets}
\begin{figure}[t]\centering%

	\begin{subfigure}[t]{.32\linewidth}
        \includegraphics[width=\textwidth]{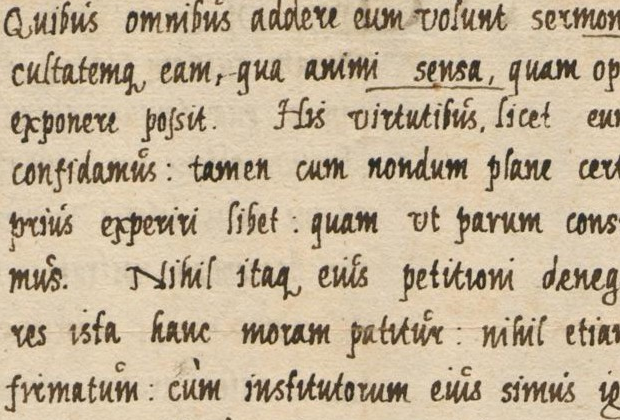}
        \caption{\icdar dataset, ID: 1379-IMG\_MAX\_309740.}
	\end{subfigure}%
	\hfill
	\begin{subfigure}[t]{.32\linewidth}
        \includegraphics[width=\textwidth]{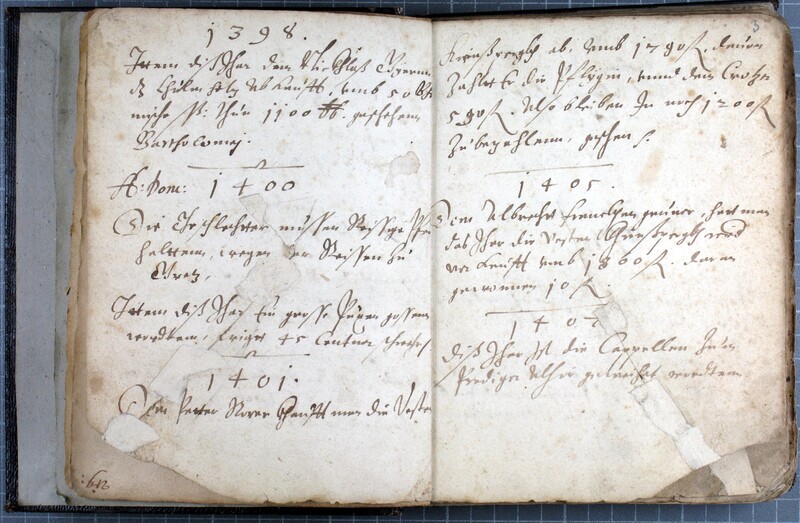}
    \caption{\czby dataset, ID: 30260559/0060}
	\end{subfigure}%
	\hfill
	\begin{subfigure}[t]{.32\linewidth}
	    \includegraphics[width=\textwidth]{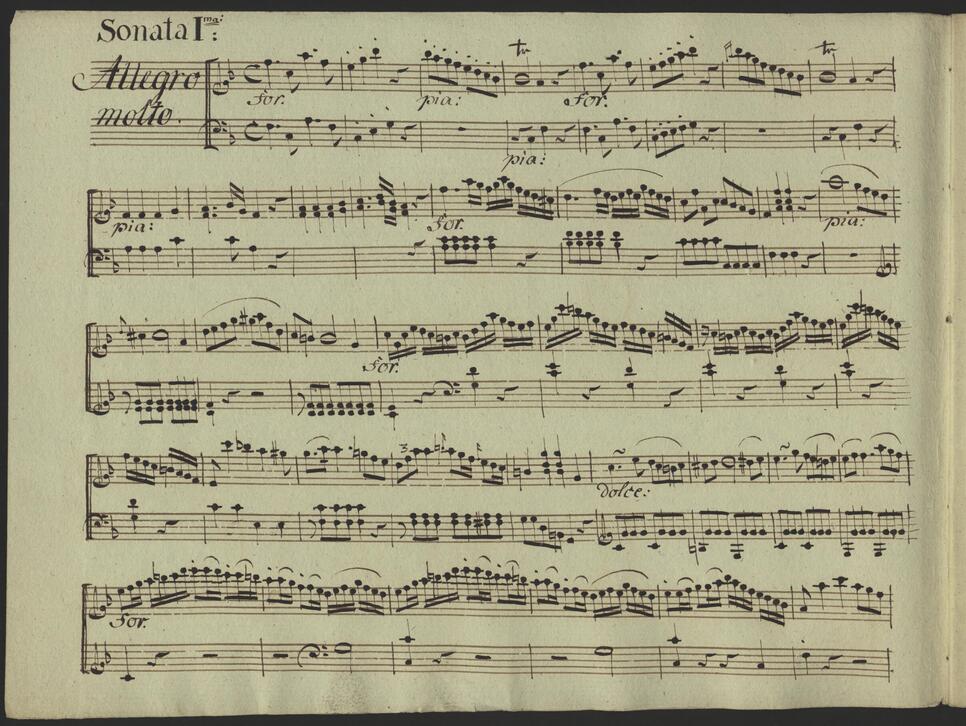}
	\caption{\music datasets, Source: SLUB, digital.slub-dresden.de/id324079575/2 (Public Domain Mark 1.0)}
	\end{subfigure}
	\caption{Examples images/excerpts of the used datasets.}
	\label{fig:db_examples}
\end{figure}

For the evaluation of different re-ranking techniques, we use three datasets: 
\begin{enumerate*}
    \item \icdar,
    \item \czby, and
    \item \music.
\end{enumerate*}
Example images can be seen in \cref{fig:db_examples}. 

\paragraph{\icdarnsc}
The \icdar dataset was proposed by \etal{Fiel}~\cite{ICDAR17WI} for the ``ICDAR 2017  competition on historical document writer identification''. 
It provides two disjunct sets, one for training and one for testing. 
The former consists of $1182$ document images, written by $394$ writers ($3$ pages per writer). 
The test set consists of $3600$ samples, written by $720$ different writers, providing $5$ pages each. 
This results in a gallery size $n_G = 5$ for the test set. 
For any test sample used as query, only four true matches remain within the test set. 
With the documents dating from \nth{13} to \nth{20} century, some of the pages had been damaged over the years, resulting in some variation of the dataset.

\paragraph{\czbynsc}
The \czby dataset\footnote{https://doi.org/10.5281/zenodo.3862591} consists of city chronicles, documents made between the \nth{15} and \nth{20} century and provided by Porta Fontium\footnote{\url{http://www.portafontium.eu}} portal. 
We used a set of 5753 images written by 143 scribes. 
Documents from 105 writers were used for testing while the remaining ones were used for training the codebook and PCA rotation matrices. 
The training set consists of \num{1496} samples while the test set contains
\num{4257} images. 
The number of samples per writer vary and range between 2 and 50 samples
(median: 44).

\paragraph{\musicnsc}
The \music dataset consists of historical hand-written music sheets of the mid
\nth{18} century provided by the SLUB
Dresden\footnote{https://www.slub-dresden.de}. They commonly do not contain much text but staves with music notes.
It contains 4381 samples, 3851 of which are used for testing coming from 35 individuals. 
The remainder of 530 samples are used for training and stems from 10 individuals. The median gallery size is $n_{G}^{median} = 40$, with a minimum gallery size of $n_{G}^{min} = 3$ and maximum $n_{G}^{min} = 660$.

\begin{figure}[t]
    \centering%
	\begin{subfigure}[t]{.49\linewidth}
		\includegraphics[width=\textwidth]{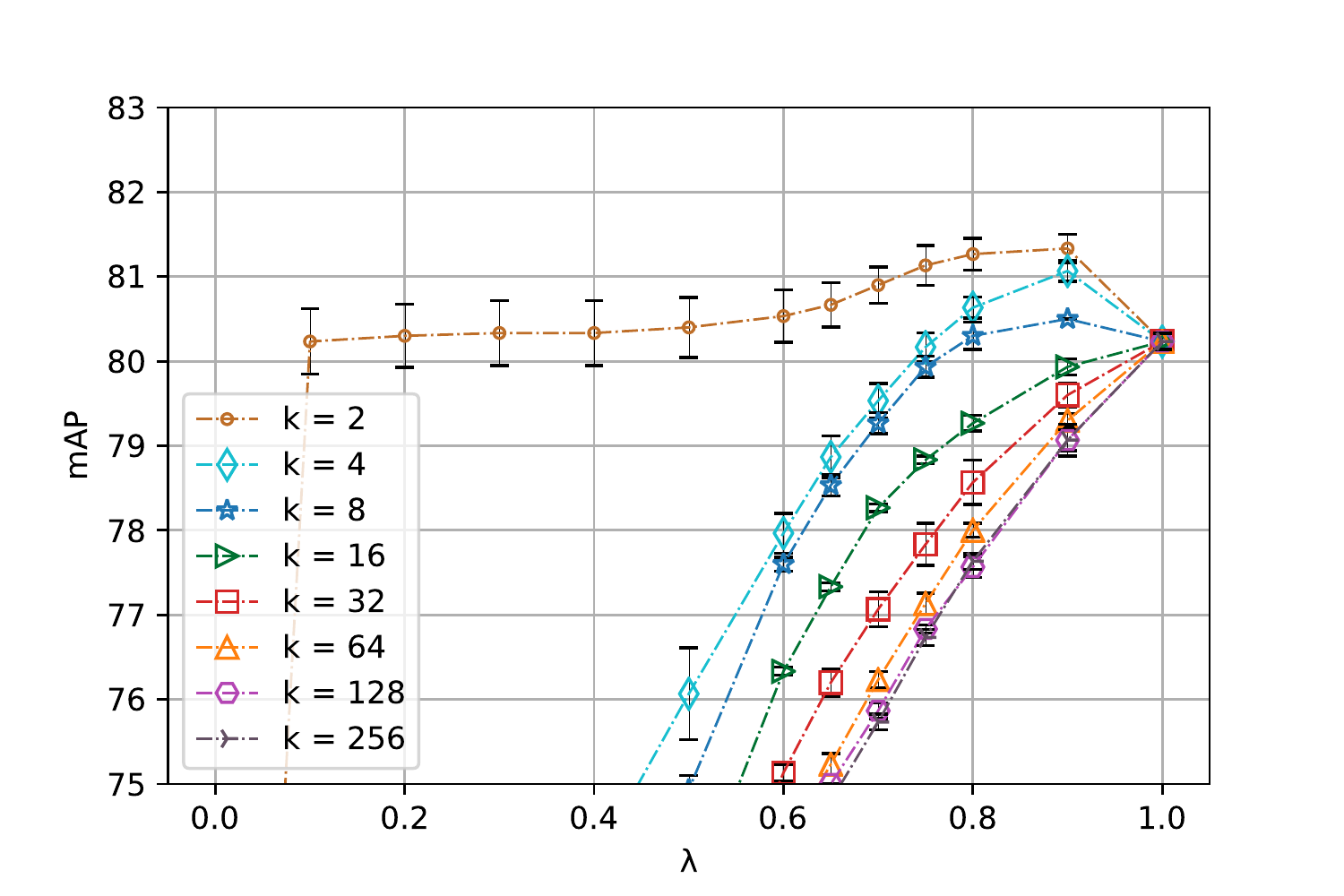}

        \caption{mAP for various combinations of $k$ and $\lambda$ performed on the \icdar training set (initial ranking achieved using cosine distances).\label{fig:icdar_train}}
        
	\end{subfigure}
	\hfill %
	\begin{subfigure}[t]{.49\linewidth}
		\includegraphics[width=\textwidth]{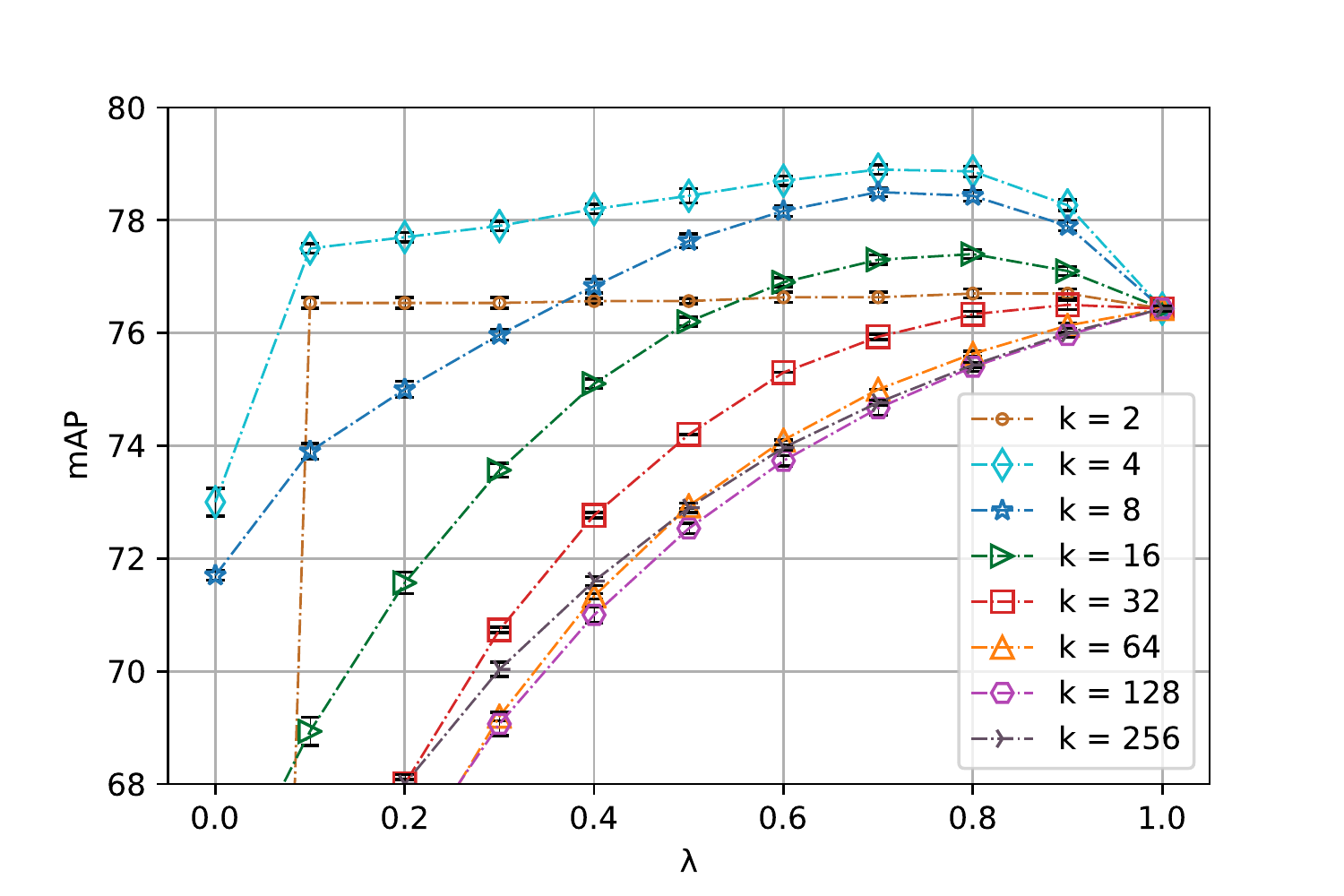}

        \caption{mAP for various combinations of $k$ and $\lambda$ performed on the \icdar test set (initial ranking achieved using ESVM-FE distances). \label{fig:icdar_test}}

	\end{subfigure}

	\caption{Parameter influence ($k$, $\lambda$) on the mAP performance of the
	Jaccard Distance method for the \icdar training and test set. For $\lambda =
1$, the distance used by this method is just the distance used for the initial
ranking. Hence for $\lambda = 1$, the performance is equal to 
the initial ranking (for all values of $k$). Regarding the test set, best results are achieved for $k = 4$, closely followed by $k = 8$.  For varying values of $k$, best results are achieved within the range $0.6 \leq \lambda \leq 0.8$. Note that the optimal parameter values ($k$, $\lambda$) differ for training and test set, which may be due to the fact that the gallery size $n_G$ is slightly higher in the test set.}
	\label{fig:softjaccard}
\end{figure}
\subsection{Results}

\subsubsection{Hyper-parameter Estimation.} Re-ranking based on $k$-reciprocal neighbors adds new hyper-parameters ($k$, $\lambda$) which need to be chosen deliberately. For this purpose, we use the training set and cosine distances. However, estimating optimal hyper-parameters using a training set builds on the assumptions that the test set is in fact represented by the data in the training set. In case of the \icdar set however, this assumption is violated because the gallery size $n_G$ differs between training and test set. In this case, the optimal values for ($k$, $\lambda$) differ between test and training set.

Note that re-ranking via the QE methods Pair- \& Triple SVM, only introduces $k$ as new hyper-parameter. However for robust estimation of $k$, we recommend to evaluate $k$ with respect to $\lambda$, even if $\lambda$ is unused.

In the following sections, we will discuss the results achieved for each dataset, individually. 

\subsubsection{Jaccard Distance Re-ranking.}

Using the initial ranking, along with the initial distances, provided by the baseline model, we are able to receive a new ranking. The new final distances $d_{\text{J} \lambda}$, \eqref{eq:finaljaccard},  depend on two parameters, $k$ for the $k$-reciprocal nearest neighbours (krNN), and $\lambda$ for the weighting between initial distance and Jaccard distance term. 

\begin{figure}[t]
    \centering%
	\begin{subfigure}[t]{.47\linewidth}
	
					\includegraphics[width=\textwidth]{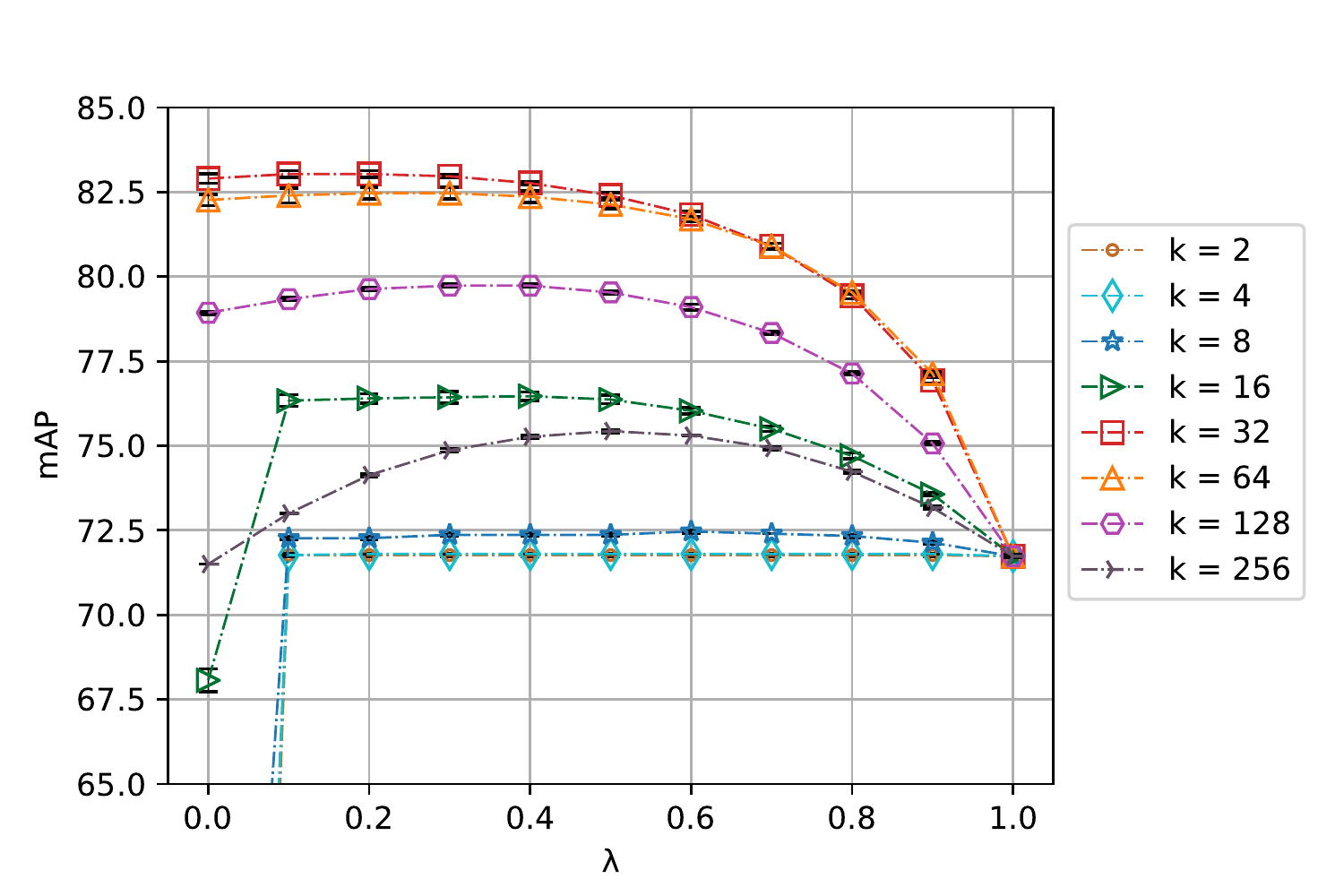}

        \caption{mAP for various combinations of $k$ and $\lambda$ performed on the \czby training set (initial ranking achieved using cosine distances).\label{fig:bycz_kl_map_train}}
        
	\end{subfigure}
	\hfill %
	\begin{subfigure}[t]{.47\linewidth}

				\includegraphics[width=\textwidth]{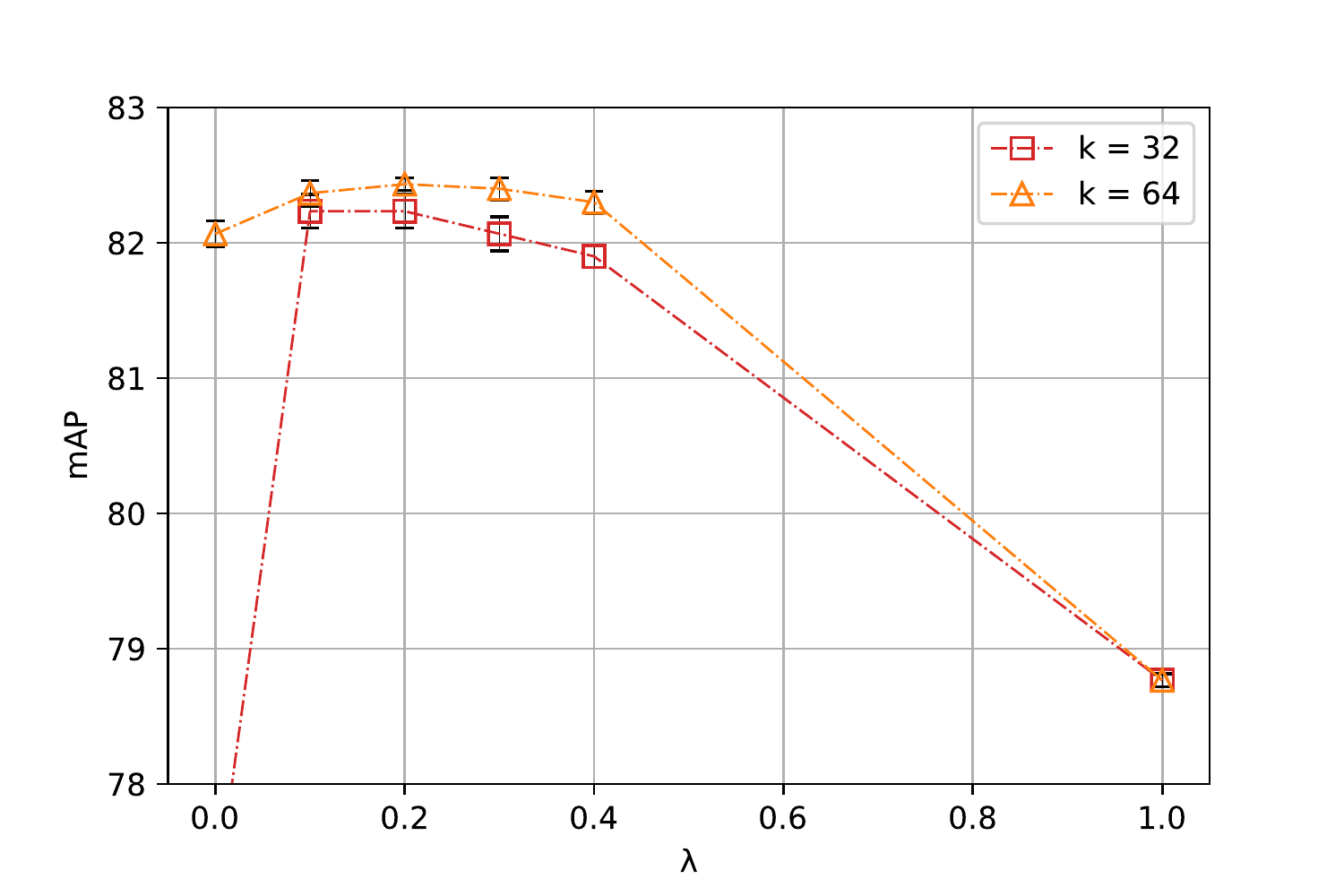}

        \caption{mAP for selected combinations of $k$ and $\lambda$ performed on the \czby test set (initial ranking achieved using ESVM-FE distances). \label{fig:bycz_kl_map_test}}

	\end{subfigure}

	\caption[\czby dataset: Parameter influence on the mAP performance and of the
	Jaccard Distance method.]{Parameter influence ($k$, $\lambda$) on the mAP
	performance of the Jaccard Distance method for the \czby training and test
set. For $\lambda = 1$, the distance used by this method is just the distance
used for the initial ranking. Hence for $\lambda = 1$, the performance is equal
to the initial ranking (for all values of $k$).}
	\label{fig:bycz_softjaccard}
\end{figure}

\paragraph{\icdar} Because the test of the \icdar dataset is not well represented by the data in the training set (the gallery size $n_G$ differs among the sets), it is of no surprise that the optimal values for ($k$, $\lambda$) differ between test and training set. To give an impression of the potential best parameters on the test set, we evaluated the influence of both parameters $k$ and $\lambda$ on both datasets, individually. Results are presented in \cref{fig:icdar_train} and \cref{fig:icdar_test} respectively.

\paragraph{\czby} In case of the \czby dataset, the gallery size $n_G$ is not
fixed but varies among the writers within $6 \leq n_G \leq 50$. 
Since the test set is drawn randomly, we can assume that the distribution of $n_G$ is roughly equal for both training and test set. Thus, we can assume that the optimal hyper-parameters ($k$, $\lambda$) for the training set will also work well on the test set. Since applying ESVM would require us to split off a `negative' set from the training set and hence reduce the training set size, we will again rely on cosine distances for hyper-parameter optimization instead of ESVM-FE. 

Various combinations of $k$ and $\lambda$ using cosine distances and the
training set are presented in \cref{fig:bycz_kl_map_train}. The optimal values
for $k$ are $k = 32$, closely followed by $k = 64$. Regarding $\lambda$, $0.1
\leq \lambda \leq 0.3$ works best. Hence we recommend using $k = 32$ and
$\lambda = 0.2$ for the \czby test set. The results are presented in
\cref{fig:bycz_kl_map_test}, along with some other selected combinations of
($k$, $\lambda$). Both $k = 32$ and $k = 64$, the best performing values for $k$
regarding the training set, were able to improve the performance on the test set
by more than \SI{3}{\percent} mAP. A drawback is the loss in Top-1 performance, which drops
by about \SI{1}{\percent}.

\begin{figure}[t]
    \centering%
	\begin{subfigure}[t]{.47\linewidth}
	
					\includegraphics[width=\textwidth]{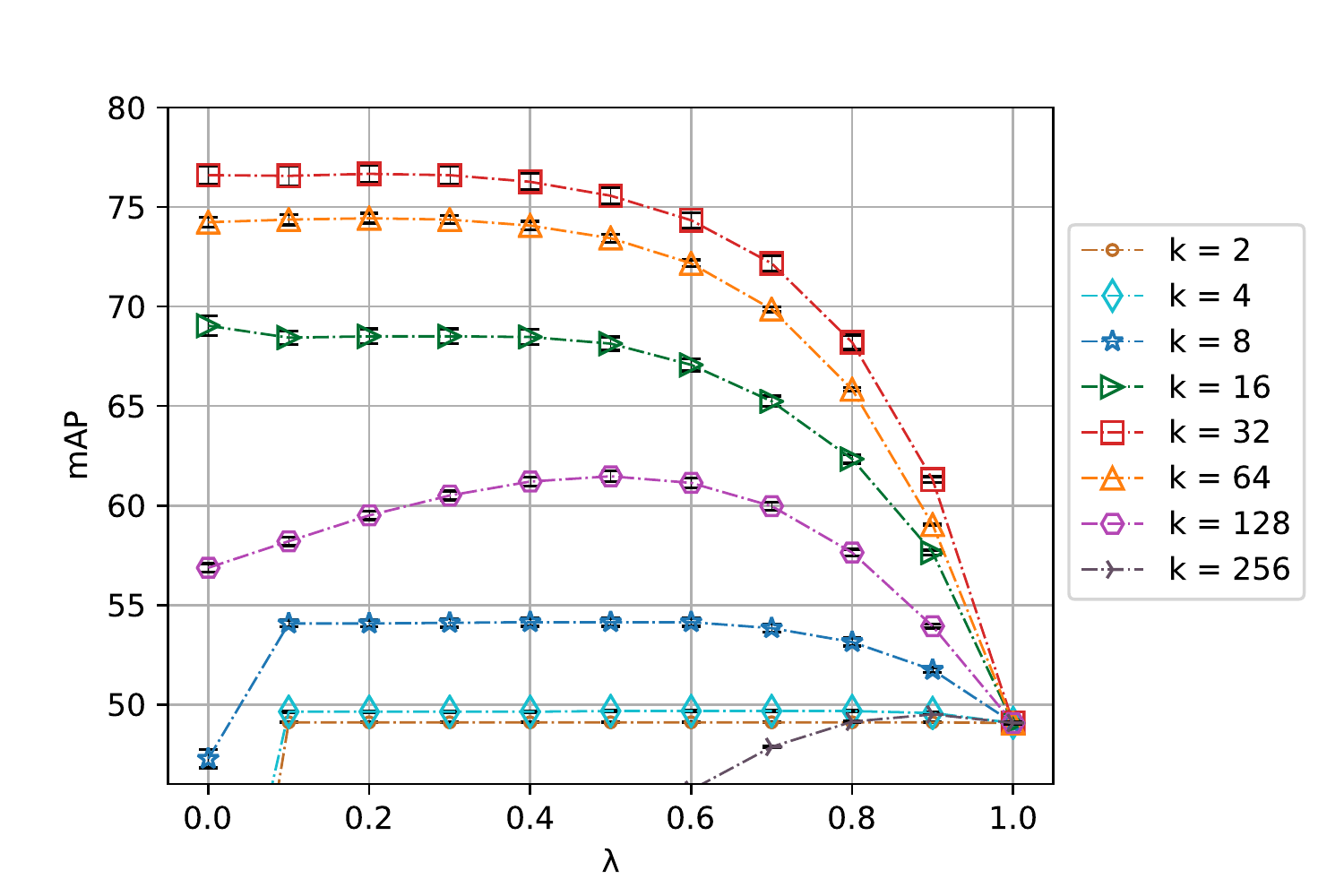}

        \caption{mAP for various combinations of $k$ and $\lambda$ performed on the \music training set (initial ranking achieved using cosine distances).\label{fig:music_kl_map_train}}
        
	\end{subfigure}
	\hfill %
	\begin{subfigure}[t]{.47\linewidth}

				\includegraphics[width=\textwidth]{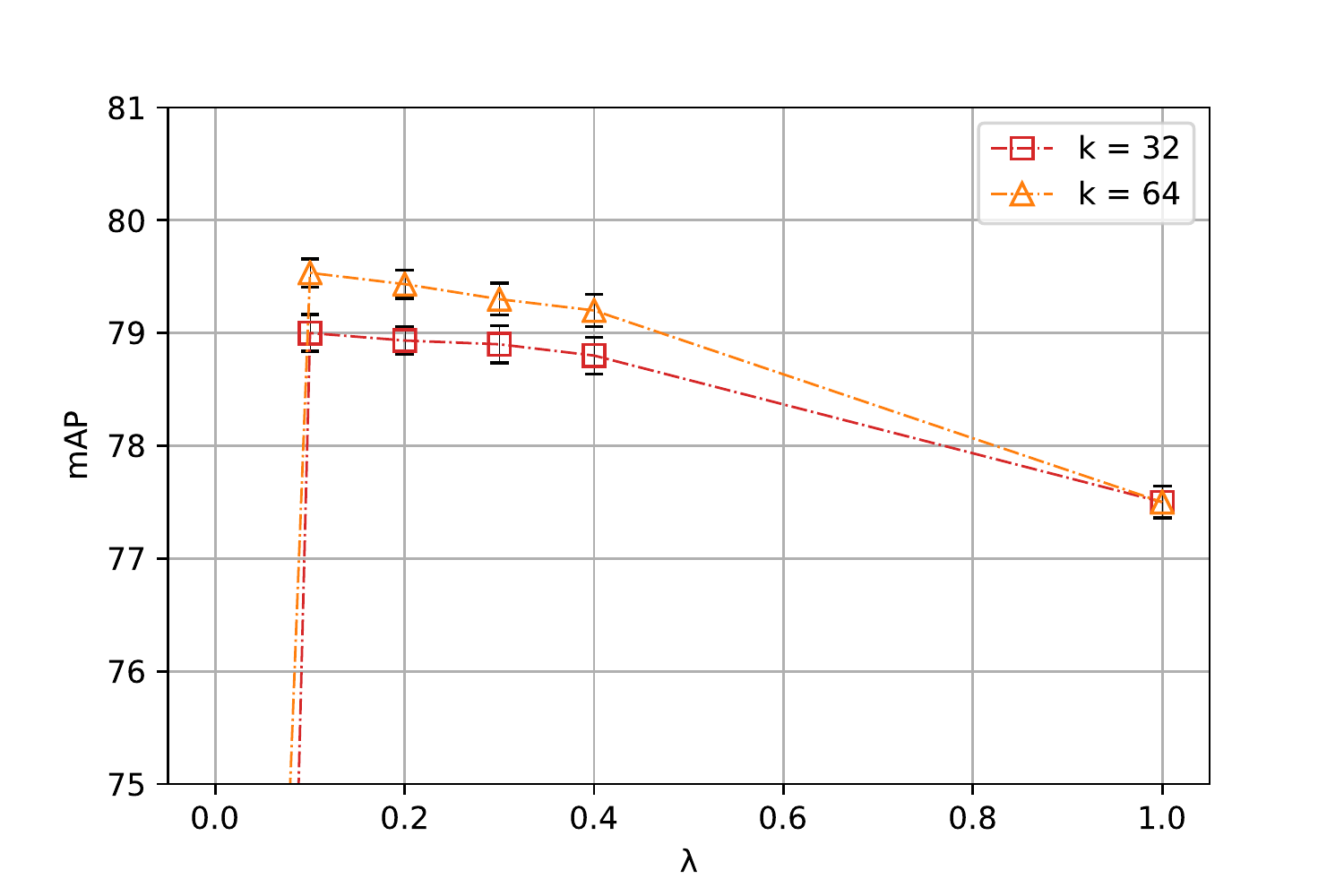}

        \caption{mAP for selected combinations of $k$ and $\lambda$ performed on the \music test set (initial ranking achieved using ESVM-FE distances). \label{fig:music_kl_map_test}}
	\end{subfigure}

	\caption{Parameter influence ($k$, $\lambda$) on the mAP performance of the
	Jaccard Distance re-ranking for the \music training and test set. For $\lambda
= 1$, the distance used by this method is just the distance used for the initial
ranking. Hence for $\lambda = 1$, the performance is equal to the initial
ranking (for all values of $k$).}
	\label{fig:music_softjaccard}
\end{figure}

\paragraph{\music} We follow the aforementioned procedure using the training set
and cosine distances to estimate the optimal hyper-parameter values ($k$,
$\lambda$) for re-ranking. Similar to the optimal values for the \czby training
set, $k = 32$ and $k = 64$ perform best, again for $\lambda = 0.2$. The results
are presented in \cref{fig:music_kl_map_train} and \cref{fig:music_kl_map_test}
showing a similar behavior than with the \czby dataset. 

\subsubsection{Pair \& Triple-SVM} As described in detail in section \ref{sec:pair_triple}, a query-expansion based re-ranking step can be added in straightforward manner by replacing the original ESVM with Pair-\&Triple-SVMs. The additional positive samples are chosen carefully by looking at the krNN, hence we use the optimal values for $k$ already estimated for the Jaccard distance method. 

\subsubsection{Comparison of Re-ranking Techniques.}
\begin{table}[t]%

\caption{Pair-\&Triple SVM methods denoted as 'krNN-P' and 'krNN-T', respectively. Baseline (ESVM-FE) also listed for comparison. For more complete results, see \Cref{tab:completeEval}.}
\label{tab:smallEval}

\begin{subtable}[c]{0.25\textwidth}
\centering
	\begin{tabular}{lcc}
		\toprule
		Method  & Top-1 & mAP \\
		\midrule
Cosine~\cite{vincentDiss}
& 83.00 & 64.03 \\

Baseline~\cite{vincentDiss}
&  89.32  &  76.43 \\

krNN-P$_{k=2}$ 
& \textbf{89.43} & \textbf{78.20} \\

krNN-T$_{k=2}$
& 89.38 & 78.19\\

Jaccard$_{k=2,\lambda=0.9}$
&  89.19  &  76.70  \\

		\bottomrule
	\end{tabular}
\caption{\icdar}
\label{tab:result_table_icdar_small}
\end{subtable}%
\hspace{4.55cm}
\begin{subtable}[c]{0.25\textwidth}
\centering
	\begin{tabular}{lcc}
		\toprule
		Method  & Top-1 & mAP \\
		\midrule
		
Cosine
& 97.72 & 77.84 \\

Baseline 
& 97.95 &  78.80 \\

krNN-P$_{k=32}$
&  98.03  &  80.06  \\

krNN-T$_{k=32}$
&  \textbf{98.04 } &  80.10 \\

Jaccard$_{k=32,\lambda=0.2}$
&  96.67  &  \textbf{82.21} \\

		\bottomrule
	\end{tabular}
\caption{\czby}
\label{tab:result_table_czby_small}
\end{subtable}

\hspace{3.7cm} %
\begin{subtable}[c]{0.25\textwidth}
\centering
	\begin{tabular}{lcc}
		\toprule
		Method  & Top-1 & mAP \\

		\midrule

Cosine
& 98.28 & 76.89  \\

Baseline 
&  98.48  &  77.49 \\

krNN-P$_{k=32}$
&  \textbf{98.62} &  78.64 \\

krNN-T$_{k=32}$
&  98.60  &  78.67 \\

Jaccard$_{k=32,\lambda=0.2}$
&  98.28  &  \textbf{78.95} \\

		\bottomrule
	\end{tabular}
\caption{\music}
\label{tab:result_table_music_small}
\end{subtable}

\end{table}

\Cref{tab:smallEval} shows a comparison between the different re-ranking techniques (more results containing Soft and Hard values can be found in \cref{tab:completeEval}, see appendix). 
All proposed re-ranking methods outperform the cosine distance-based ranking (cosine) and the ranking based on the ESVM feature transform (baseline) in terms of mAP significantly. 
Jaccard distance re-ranking is mAP-wise the best performing re-ranking technique.
While the results show that the overall ranking gets improved, the Top-1 accuracy using the Jaccard distance re-ranking gets affected negatively. 
In case of Pair/Triplet-SVM this is not the case and can therefore be seen as a good compromise. 

\section{Conclusion}
In this paper, we investigated the benefits of including a re-ranking step to a writer retrieval pipeline.
We presented two ways of re-ranking, one based on the Jaccard distance and another one on an improved query expansion. 
We show that both methods achieve significantly higher mAP than the state-of-the-art baseline.
While the Jaccard distance re-ranking returns the highest retrieval rates, the Top-1 accuracy drops. 
This makes a usage of Pair-SVM more suitable in case of identification.

Our results strongly hint that the gallery size has an impact on the optimal value for the number of reciprocal neighbors $k$.
For larger galleries, larger $k$ values are better, and the re-ranking is more efficient.
However, even for smaller amounts of data, such as the \icdar dataset, using one of the proposed re-ranking approaches improves the current state-of-the-art results.

There is however still room for improvement, and further investigations.
We believe that the proposed method could benefit from the assignment of weights to the friend samples in the positive set in order to attribute them different importance based on their ranking. 
The two different techniques, Pair/Triple-SVMs and Jaccard distance re-ranking, could potentially also be combined to improve the results further.

\section*{Acknowledgement}
This work has been partly supported by the Cross-border Cooperation Program Czech Republic -- Free State of Bavaria ETS Objective 2014-2020 (project no.\ 211).

\bibliographystyle{splncs04}
\bibliography{references}

\begin{thebibliography}{10}
\providecommand{\url}[1]{\texttt{#1}}
\providecommand{\urlprefix}{URL }
\providecommand{\doi}[1]{https://doi.org/#1}

\bibitem{arandjelovic12three}
Arandjelovi{\'c}, R., Zisserman, A.: Three things everyone should know to
  improve object retrieval. In: 2012 IEEE Conference on Computer Vision and
  Pattern Recognition. pp. 2911--2918. IEEE (2012)

\bibitem{Brink12}
Brink, A., Smit, J., Bulacu, M., Schomaker, L.: Writer identification using
  directional ink-trace width measurements. Pattern Recognition
  \textbf{45}(1),  162--171 (jan 2012)

\bibitem{Bulacu07}
Bulacu, M., Schomaker, L.: Automatic handwriting identification on medieval
  documents. In: 14th International Conference on Image Analysis and Processing
  (ICIAP 2007). pp. 279--284. Modena (sep 2007)

\bibitem{Canny}
Canny, J.: A computational approach to edge detection. IEEE Transactions on
  Pattern Analysis and Machine Intelligence  \textbf{8}(6),  679--698 (nov
  1986)

\bibitem{carpineto12survey}
Carpineto, C., Romano, G.: A survey of automatic query expansion in information
  retrieval. ACM Computing Surveys (CSUR)  \textbf{44}(1), ~1 (2012)

\bibitem{vincentDiss}
Christlein, V.: Handwriting Analysis with Focus on Writer Identification and
  Writer Retrieval. doctoralthesis, Friedrich-Alexander-Universit{\"a}t
  Erlangen-N{\"u}rnberg (FAU) (2019)

\bibitem{christlein17writer}
Christlein, V., Bernecker, D., H{\"o}nig, F., Maier, A., Angelopoulou, E.:
  Writer identification using gmm supervectors and exemplar-svms. Pattern
  Recognition  \textbf{63},  258--267 (2017)

\bibitem{Christlein15GCPR}
Christlein, V., Bernecker, D., Maier, A., Angelopoulou, E.: Offline writer
  identification using convolutional neural network activation features. In:
  Gall, J., Gehler, P., Leibe, B. (eds.) Pattern Recognition: 37th German
  Conference, GCPR 2015, Aachen, Germany, October 7-10, 2015, Proceedings,
  vol.~9358, pp. 540--552. Springer International Publishing (2015)

\bibitem{christlein17unsup}
Christlein, V., Gropp, M., Fiel, S., Maier, A.: Unsupervised feature learning
  for writer identification and writer retrieval. In: 2017 14th IAPR
  International Conference on Document Analysis and Recognition (ICDAR).
  vol.~1, pp. 991--997. IEEE (2017)

\bibitem{Christlein18DAS}
Christlein, V., Maier, A.: {Encoding CNN Activations for Writer Recognition}.
  In: 13th IAPR International Workshop on Document Analysis Systems. pp.
  169----174. Vienna (apr 2018)

\bibitem{chum11total}
Chum, O., Mikulik, A., Perdoch, M., Matas, J.: Total recall ii: Query expansion
  revisited. In: CVPR 2011. pp. 889--896. IEEE (2011)

\bibitem{chum07total}
Chum, O., Philbin, J., Sivic, J., Isard, M., Zisserman, A.: Total recall:
  Automatic query expansion with a generative feature model for object
  retrieval. In: 2007 IEEE 11th International Conference on Computer Vision.
  pp.~1--8. IEEE (2007)

\bibitem{ICDAR17WI}
Fiel, S., Kleber, F., Diem, M., Christlein, V., Louloudis, G., Nikos, S.,
  Gatos, B.: Icdar2017 competition on historical document writer identification
  (historical-wi). In: 2017 14th IAPR International Conference on Document
  Analysis and Recognition (ICDAR). vol.~01, pp. 1377--1382 (Nov 2018)

\bibitem{Fiel13}
Fiel, S., Sablatnig, R.: Writer identification and writer retrieval using the
  fisher vector on visual vocabularies. In: 2013 12th International Conference
  on Document Analysis and Recognition (ICDAR). pp. 545--549. Washington DC
  (aug 2013)

\bibitem{Jegou12ALI}
J{\'{e}}gou, H., Perronnin, F., Douze, M., S{\'{a}}nchez, J., P{\'{e}}rez, P.,
  Schmid, C.: Aggregating local image descriptors into compact codes. IEEE
  Transactions on Pattern Analysis and Machine Intelligence  \textbf{34}(9),
  1704--1716 (sep 2012)

\bibitem{Keglevic18}
Keglevic, M., Fiel, S., Sablatnig, R.: Learning features for writer retrieval
  and identification using triplet cnns. In: 2018 16th International Conference
  on Frontiers in Handwriting Recognition. pp. 211--216. Niagara Falls (aug
  2018)

\bibitem{Kobayashi14}
Kobayashi, T.: Dirichlet-based histogram feature transform for image
  classification. In: 2014 IEEE Conference on Computer Vision and Pattern
  Recognition (CVPR). pp. 3278--3285. Columbus (jun 2014)

\bibitem{lowe04distinctive}
Lowe, D.G.: Distinctive image features from scale-invariant keypoints.
  International journal of computer vision  \textbf{60}(2),  91--110 (2004)

\bibitem{Murray16}
Murray, N., Jegou, H., Perronnin, F., Zisserman, A.: Interferences in match
  kernels. IEEE Transactions on Pattern Analysis and Machine Intelligence
  \textbf{39}(9),  1797--1810 (oct 2016)

\bibitem{Perronnin10IFK}
Perronnin, F., S{\'{a}}nchez, J., Mensink, T.: Improving the fisher kernel for
  large-scale image classification. In: Daniilidis, K., Maragos, P., Paragios,
  N. (eds.) Computer Vision - ECCV 2010: 11th European Conference on Computer
  Vision, Heraklion, Crete, Greece, September 5-11, 2010, Proceedings, Part IV,
  pp. 143--156. Lecture Notes in Computer Science, Springer, Berlin, Heidelberg
  (2010)

\bibitem{qin11hello}
Qin, D., Gammeter, S., Bossard, L., Quack, T., Van~Gool, L.: Hello neighbor:
  Accurate object retrieval with k-reciprocal nearest neighbors. In: CVPR 2011.
  pp. 777--784. IEEE (2011)

\bibitem{Tang16}
Tang, Y., Wu, X.: Text-independent writer identification via cnn features and
  joint bayesian. In: 2016 15th International Conference on Frontiers in
  Handwriting Recognition (ICFHR). pp. 566--571. Shenzhen (oct 2016)

\bibitem{Wahlberg15}
Wahlberg, F., M{\aa}rtensson, L., Brun, A.: Large scale style based dating of
  medieval manuscripts. In: 3rd International Workshop on Historical Document
  Imaging and Processing (HIP'15). pp. 107--114. ACM, Nancy (aug 2015)

\bibitem{zhong17re}
Zhong, Z., Zheng, L., Cao, D., Li, S.: Re-ranking person re-identification with
  k-reciprocal encoding. In: Proceedings of the IEEE Conference on Computer
  Vision and Pattern Recognition. pp. 1318--1327 (2017)

\bibitem{Zhou10NGV}
Zhou, X., Zhuang, X., Tang, H., Hasegawa-Johnson, M., Huang, T.S.: Novel
  gaussianized vector representation for improved natural scene categorization.
  Pattern Recognition Letters  \textbf{31}(8),  702--708 (jun 2010)

\end{thebibliography}

\pagebreak
\section*{Appendix}
\begin{table}[htbp]%\small
\setlength{\aboverulesep}{0pt}
\setlength{\belowrulesep}{0pt}
\setlength{\extrarowheight}{.55ex}

\caption{Full results. Pair-\&Triple SVM methods denoted as 'krNN-P' and 'krNN-T' respectively. Baseline (ESVM-FE) also listed for comparison.}
\label{tab:completeEval}

\begin{subtable}[c]{\textwidth}
\centering
	\begin{tabular}{lccccccc}
		\toprule
		Method  & Top-1 & Hard-2 & Hard-3 & Hard-4 &  Soft-5 & Soft-10 & mAP \\

		\midrule
Cosine
& 83.00  & 66.86  & 50.11  & 30.59  & 87.92  & 89.43  & 64.03 \ $\pm$0.29\\

Baseline 
&  89.32  &  79.09  &  67.68 &  48.74  &   \textbf{92.96}  &  93.83 &  76.43 \ $\pm$0.06\\

krNN-P$_{k=2}$
& 89.43  & \textbf{81.68}  & 71.83  & 54.49  & 92.06  & 93.31  & 78.20 \ $\pm$0.14  \\

\textcolor{gray}{krNN-P$_{k=4}$}
&  \textbf{89.55}  &  81.49  &  71.89  &  \textbf{57.78}  &  91.92  &  92.95  &  78.80 \ $\pm$0.25  \\

krNN-T$_{k=2}$
& 89.38  & 81.63  & 71.79  & 54.44  & 92.06  & 93.30  & 78.19 \ $\pm$0.16 \\

\textcolor{gray}{krNN-T$_{k=4}$}
&  89.52  &  81.25 & 71.61 & 57.51 & 91.95 & 93.01 & 78.75 \ $\pm$0.26 \\

Jaccard$_{k=2,\lambda=0.9}$
&  89.19  &  80.33  &  68.25  &  49.03  &  92.95  &  93.83  &  76.70 \ $\pm$0.01 \\

\textcolor{gray}{Jaccard$_{k=4,\lambda=0.7}$}
&  88.47  &  81.48  &  \textbf{73.35}  &  57.49  &  92.71 &  \textbf{93.84}  &  \textbf{78.91} $\pm$0.10 \\
		\bottomrule
	\end{tabular}
\caption{\icdar}
\label{tab:result_table_icdar}
\end{subtable}%
% \end{table}
% \begin{table}[htbp]
% \ContinuedFloat

\begin{subtable}[c]{\textwidth}
\centering
	\begin{tabular}{lccccccc}
		\toprule
		Method  & Top-1 & Hard-2 & Hard-3 & Hard-4 &  Soft-5 & Soft-10 & mAP \\
		\midrule

Cosine
& 97.72  & 95.46  & 93.53  & 91.77  & 98.64  & 98.90  & 77.84 \ $\pm$0.04 \\
		
Baseline 
 & 97.95  & 95.69  & 94.00  & 92.07  & \textbf{98.80}  & \textbf{98.99 } & 78.80 \ $\pm$0.14\\

krNN-P$_{k=32}$
 & 98.03  & \textbf{96.47}  & \textbf{94.73 } &\textbf{ 93.20}  & 98.61  & 98.80  & 80.06 \ $\pm$0.20 \\

krNN-T$_{k=32}$
 & \textbf{98.04}  & 96.44  & 94.61  & 93.11  & 98.60  & 98.80  & 80.10 \ $\pm$0.24  \\

Jaccard$_{k=32,\lambda=0.2}$
& 96.67  & 94.88  & 93.37  & 91.93  & 97.83  & 98.30  & \textbf{82.21} \ $\pm$0.06 \\

		\bottomrule
	\end{tabular}
\caption{\czby}
\label{tab:result_table_czby}
\end{subtable}

% \end{table}
% \begin{table}[htbp]
% \ContinuedFloat
\begin{subtable}[c]{\textwidth}
\centering
	\begin{tabular}{lccccccc}
		\toprule
		Method  & Top-1 & Hard-2 & Hard-3 & Hard-4 &  Soft-5 & Soft-10 & mAP \\

		\midrule

Cosine
& 98.28  & 97.43  & 96.82  & 95.94  & 99.12  & 99.21  & 76.89 \ $\pm$0.30  \\

Baseline 
&  98.48  &  97.69  &  96.92  &  96.10  &  \textbf{99.15}  &  99.31  &  77.49 \ $\pm$0.15 \\

krNN-P$_{k=32}$
&  \textbf{98.62}  &  \textbf{98.04}  &  97.67  &  96.86  &  99.13  &  \textbf{99.32}  &  78.64 \ $\pm$0.11 \\

krNN-T$_{k=32}$
&  98.60  &  98.03  &  \textbf{97.70}  &  \textbf{96.89}  &  99.13  &  99.31  &  78.67 \ $\pm$0.14 \\

Jaccard$_{k=32,\lambda=0.2}$
&  98.28  &  97.45  &  97.01  &  96.39  &  99.00  &  99.17  &  \textbf{78.95} \ $\pm$0.03 \\

		\bottomrule
	\end{tabular}
\caption{\music}
\label{tab:result_table_music}
\end{subtable}

\end{table}

\end{document}